\documentclass[twoside]{article}

\usepackage[accepted]{aistats2023_my}
\usepackage[round]{natbib}

\usepackage{hyperref}       
\usepackage{amsmath}
\usepackage{amssymb}
\usepackage{graphicx}
\usepackage{wrapfig}
\usepackage{tikz}
\usepackage{stmaryrd} 	

\usepackage{bm}
\newcommand{\bk}{\bm{k}}
\newcommand{\bx}{\bm{x}}
\newcommand{\ff}{\bm{f}}   
\newcommand{\bxi}{\bm{\xi}}
\newcommand{\bnu}{\bm{\nu}}
\newcommand{\bkk}{\bm{\kappa}} 	
\newcommand{\bb}{\bm{b}} 		
\newcommand{\bh}{\bm{h}} 		
\newcommand{\by}{\bm{y}} 		
\newcommand{\etac}{\lceil\eta\rceil} 		

\newcommand{\dd}{\;\mathrm{d}}
\newcommand{\R}{\mathbb{R}}
\newcommand{\cF}{\mathcal{F}}
\newcommand{\cS}{\mathcal{S}}

\makeatletter
\newcommand{\xRightarrow}[2][]{\ext@arrow 0359\Rightarrowfill@{#1}{#2}}
\makeatother

\begin{document}

%

%

\twocolumn[

\aistatstitle{Scale invariant process regression: Towards Bayesian ML with minimal assumptions}

\aistatsauthor{ Matthias Wieler }

\aistatsaddress{ Bosch Center for Artificial Intelligence (BCAI) \\ Oct. 30, 2022} ]

\begin{abstract}

Current methods for regularization in machine learning require quite specific model assumptions (e.g.\ a kernel shape) that are not derived from prior knowledge about the application, but must be imposed merely to make the method work.
We show in this paper that regularization can indeed be achieved by assuming nothing but invariance principles (w.r.t.\ scaling, translation, and rotation of input and output space) and the degree of differentiability of the true function.
Concretely, we derive a novel (non-Gaussian) stochastic process from the above minimal assumptions, and we present a corresponding Bayesian inference method for regression.
The mean posterior turns out to be a polyharmonic spline, and the posterior process is a mixture of $t$-processes.
Compared with Gaussian process regression, the proposed method shows equal performance and has the advantages of being (i) less arbitrary (no choice of kernel) (ii) potentially faster (no kernel parameter optimization), and (iii) having better extrapolation behavior. 
We believe that the proposed theory has central importance for the conceptual foundations of regularization and machine learning and also has great potential to enable practical advances in ML areas beyond regression.


\end{abstract}

\section{Introduction}

\paragraph{Minimal assumptions for regularization}

A fundamental issue in machine learning is that good generalization can only be achieved with regularization, which requires a smoothness assumption.
We want to point out that the terms ``regularization'' and ``smoothness assumption'' are in fact quite inappropriate, because currently available regularizing models do not only assume smoothness or regularity in the sense of differentiability, but make much stronger assumptions, e.g.\ setting the value of a regularization-controlling parameter, restricting the number of model parameters, fixing the length scale, choosing the kernel shape, etc.
Assuming differentiability alone seems to be insufficient, because even infinitely often differentiable functions can be arbitrarily complex on an arbitrarily short scale.
Therefore, ``regularization'' actually means simplification, and ``smoothness assumptions'' are in fact simplicity assumptions.

The motivation for the present work is to derive a regularization method from as little assumptions as possible.
We find that it is indeed sufficient to assume regularity, i.e.\ a certain degree of differentiability of the true function.
In this sense our work justifies the terms ``regularization'' and ``smoothness assumption'' in retrospect.

The essential ingredient to achieve the above is scale invariance, and we found that the most suitable setting to carry out our approach is regression\footnote{opposed to classification, density estimation, clustering etc.}.
In this setting, we assume translation-, scale-, and rotation invariance in input and output space.
We want to mention that from an objective Bayesian viewpoint, these assumptions are justified by the prior knowledge that input and output space have Euclidean structure.
The absence of a distinguished position, scale, or orientation already requires corresponding invariances.
Since a large number of applications have Euclidean input and output space, the above invariance assumptions can be said to characterize the standard setting of regression.

There is one more assumption required by our approach, which will be discussed in Sec.\ \ref{sec:SI_distr}.


\paragraph{Scale invariance and Gaussian processes}

Gaussian processes are the currently leading 
method for regression and are therefore our methodical reference and performance benchmark.
Scale invariance appears in Gaussian process regression in connection with the Wiener process (and related processes) and with splines.

The Wiener process is self-similar in the sense that if $W(x)$ is a Wiener process, then $W(sx)/\sqrt{s}$ also is a Wiener process. 
Self-similarity is sometimes called ``scale invariance'', but we want to stress that the Wiener process is invariant only under a \emph{simultaneous} change of $x$- and $y$-scale, whereas the proposed scale invariant process (SIP) is invariant under a change of either $x$- or $y$-scale independently of each other.

Still, there is a close relation between SIP and the Wiener process in that both processes have a power-law spectrum.
As a consequence, the maximum posterior can be described by minimizing a functional involving differential operators, which results in splines.
The relation between Gaussian processes (e.g.\ the Wiener process), differential operators, and splines is well-known, see \citet{rasmussen_gaussian_2005}, \citet{kimeldorf_correspondence_1970}, \citet{dold_splines_1977}, \citet{wahba_spline_1990}, \citet{kent_link_1994}.
We encounter the same theory from a somewhat different perspective and extend it to fractional differentials, which is quite straightforward from a frequency space perspective.


One drawback of self-similar GPs is that they are not stationary, and in practice, almost exclusively stationary GPs are used.
Therefore, we will mostly refer to stationary GPs in the Results and Discussion sections, although they are theoretically less closely related to the scale invariant process.

\paragraph{Overview of paper}

Sec.\ 2 defines the scale invariant process, Sec.\ 3 derives the posterior process for interpolation (no observation error), and Sec.\ 4 describes the regression method.
To span the arc from a novel stochastic process to a practically working method, the exposition is quite condensed.
Sec.\ 5 gives some exemplary results, and Sec.\ 6 discusses the most important implications.

\section{Scale invariant process (SIP)}

To give a first impression of the scale invariant process, we mention that for a single input dimension, it is very similar\footnote{equivalence remains to be investigated}
to a hierarchical model involving the $l$-times integrated Wiener process $W^{(-l)}$, where the hyperparameters $u_x,u_y,h_x,h_y$ have been incorporated into the stochastic process: 
\begin{align}
	f(x) &= u_y + h_y\cdot W^{\left(\frac{1}{2}-\eta\right)}(u_x + h_x x) \label{hierarchical_GP}
\end{align}
\begin{align}
	p(u_{x,y}) &\propto 1 && u_{x,y}\in\R \\
	p(h_{x,y}) &\propto \frac{1}{h_{x,y}} && h_{x,y}\in\R_+
\end{align}
The following construction starts at a very different point, however: by generalizing a finite-dimensional scale invariant distribution to a scale invariant process.

\subsection{Scale invariance in $\nu$ dimensions}
\label{sec:SI_distr}

We consider a distribution over $\nu$ function values $\ff=[f_1,\dots,f_\nu] = [f(\bx_1),\dots,f(\bx_\nu)]$.
It is scale invariant iff is satisfies
\begin{align}
	p(\ff) \dd\ff &\;=\; p(s\ff) \,s^\nu \dd\ff \\
	p(s\ff) &\;=\; \frac{1}{s^\nu} \, p(\ff) \label{radial_SI}
\end{align}
While the functional equation (\ref{radial_SI}) determines the radial behavior of $p(\ff)$, the tangential behavior is left unspecified.
We would like to apply the maximum entropy principle to choose the ``simplest'' or ``most uniform'' distribution, but to our knowledge, there is unfortunately no valid definition of entropy for improper distributions.
However, it seems quite obvious that a spherically symmetric distribution would be the simplest choice in this case.
For our purposes, we need different length scales $a_i$ in different dimensions (see Sec.\ \ref{sec:SIP_spectrum}), and the corresponding ``obvious'' simplest choice is to assume ellipsoidal symmetry.
\begin{align}
	p_\text{SI}(\ff) &= \frac{1}{\|\ff\|_a^\nu} \label{SI_nudim} \\
	\|\ff\|_a^2 &= \sum_{i=1}^\nu \left(\frac{f_i}{a_i}\right)^2 \label{ellipsoidal_norm}
\end{align}
The coefficients $a_i\in\R_+$ can be interpreted as amplitudes and will be specified below.
Note that the distribution (\ref{SI_nudim}) is improper ``in both directions'', i.e.\ its integral diverges for both $\|\ff\|\rightarrow0$ and $\|\ff\|\rightarrow\infty$.

\subsection{Scale invariant process}
\label{sec:SI_process}

We now consider real-valued functions $f(\bx)$ in Euclidean space $\bx\in\R^D$ and define the scale invariant process such that for any finite set $\{\bx_1,\dots,\bx_\nu\}$ the marginal distribution is described by (\ref{SI_nudim}).
According to the Daniell--Kolmogorov theorem, such a process exists if the collection of marginals satisfies the following consistency requirement for all disjunct $\ff_1,\ff_2$:
\begin{equation}
	p(\ff_1) = \int p(\ff_1,\ff_2) \dd\ff_2 \label{marginal_consistency}
\end{equation}
For $p(\ff)$ defined by (\ref{SI_nudim}), this equation holds for any choice 
of $a_i$, as we will show in the Appendix.

Note that the scale invariant process is improper, just like the scale invariant distributions.
We are not aware of any theory about improper processes, but since we did not encounter unexpected problems, we are confident that the construction is indeed valid and well-defined.

\subsection{Spectrum}
\label{sec:SIP_spectrum}

From now on we work in frequency space, which ensures stationarity of the process and introduces correlations between different input points $\bx_i$ in a natural way.
Scale invariance in $\bx$ requires that no scale (i.e.\ no frequency $\bk$) is distinguished, which is the case only for power-law spectra.
\begin{align}
	a_\eta(\bk) &= \frac{1}{\|\bk\|^{\eta+\frac{D}{2}}} \qquad \eta\in(\R_+\!\setminus\!\mathbb{N}_+) \label{eta_spectrum}
\end{align}
The parameter $\eta$ is discussed below.
The continuous version of (\ref{ellipsoidal_norm}) now reads
\begin{align}
	|f|_\eta^2 &= \int_{\R^D} \left|\frac{\hat f(\bk)}{a_\eta(\bk)}\right|^2 \dd\bk \label{eta_norm}
\end{align}
where $\hat f(\bk)=\hat f^*(-\bk)$ is the Fourier transform of $f(\bx)$.


\subsection{Regularity}

The regularity of the process's sample functions $f(\bx)$ is determined by the asymptotic behavior of $a_\eta(\bk)$ for $\|\bk\|\rightarrow\infty$.
The form of the exponent in (\ref{eta_spectrum}) is chosen such that $\eta$ describes the degree of differentiability. 
More exactly, $f(\bx)$ is almost surely $C$ times continuously differentiable, with the highest existing derivative being $\alpha$-H\"older continuous for all $\alpha<H$, where $C$ and $H$ are given by:
\begin{align}
	&\text{degree of differentiability:}& C &= \lfloor\eta\rfloor \label{differentiability} \\
	&\text{H\"older/Hurst exponent:}& H &= \eta - \lfloor\eta\rfloor \label{Hurst}
\end{align}
The transition between different degrees of differentiability happens at integer $\eta$, which poses special mathematical problems, and we therefore exclude these cases throughout this paper.
The H\"older exponent is usually called ``Hurst exponent'' in the context of fractional Brownian motion.

For Gaussian processes, some special cases of (\ref{differentiability}, \ref{Hurst}) have long been known.
\citet{flandrin_spectrum_1989} gives the result for fractional Brownian motion ($D=1$, $\eta\in(0,1)$), and \citet{molchan_problems_1967} gives the result for Lévy $D$-parameter Brownian motion ($D\in\mathbb{N}_+$, $\eta=1/2$).
As we show in the Appendix, this result also holds for the scale invariant process for all combinations of $(D,\eta)$.

Note that the exponent of (\ref{eta_spectrum}) also appears in the power spectrum of the Matèrn kernel (length-scale denoted by $l$):
\begin{align}
	S_\text{Matèrn}(\bk) = a_\text{Matèrn}^2(\bk)& \propto \left(\!\frac{\eta}{2\pi^2 l^2 + \|\bk\|^2} \!\right)^{\!\!-\left(\eta + \!\frac{D}{2}\!\right)} \\
	a_\text{Matèrn}(\bk)& \xrightarrow{\|\bk\|\rightarrow\infty} \|\bk\|^{-\left(\eta+\frac{D}{2}\right)}
\end{align}
The regularity parameter $\eta$ is usually denoted $\nu$ in the context of Matèrn kernels and chosen as a half-integer $1/2$, $3/2$, or $5/2$, which is well known to result in sample paths that are differentiable 0, 1, and 2 times, resp.

\subsection{Nullspace}


Th $\eta$-norm (\ref{eta_norm}) vanishes for all polynomials up to degree $\lfloor\eta\rfloor$, or more exactly:
\begin{equation}
	|\bx^{\bnu}|_\eta = \left|\prod_{d=1}^D x_d^{\nu_d}\right|_\eta =
	\begin{cases}
		0 & |\bnu|<\eta \\
		\infty & |\bnu|>\eta
	\end{cases}
\end{equation}
where $\bnu=[\nu_1,\dots,\nu_D]$ is a multi-index with absolute value $|\bnu| = \sum_d \nu_d$.
This result can be found e.g.\ in \citep{wahba_spline_1990}, and we rederive it in the Appendix.
The number of vanishing monomials, i.e.\ the dimension of the nullspace $\mathcal{N}_\eta$ can be expressed as the binomial coefficient
\begin{equation}
	N_0 = \binom{\lfloor\eta\rfloor + D}{\lfloor\eta\rfloor} \label{N0}
\end{equation}


\subsection{Corresponding function space}

To deal with the scale invariant process, it is convenient to use the function space $\cF_\eta$ that is defined by the inner product
\begin{equation}
	\langle f,g\rangle_\eta = \int_{\R^D} \frac{\hat f(\bk) \, \hat g^*(\bk)}{a_\eta^2(\bk)} \dd\bk \label{inner_product}
\end{equation}
This inner product induces the $\eta$-norm (\ref{eta_norm}) and as  result, the scale invariant process is isotropic in $\cF_\eta$.
Note that due to the nullspace, $\cF_\eta$ is \emph{not} a Hilbert space and in particular not an RKHS.


In this context, we mention that to deal with such nullspaces from an RKHS perspective, the concepts of ``intrinsic random function'' and ``generalized covariance'' have been introduced \citep{rasmussen_gaussian_2005}, \citep{cressie_statistics_1993}, \citep{stein_interpolation_1999}.
However, we have found no necessity for such additional concepts in our treatment.

\section{Interpolation}
\label{sec:Interpolation}

\begin{figure}
	\centering
	\includegraphics[width=\linewidth]{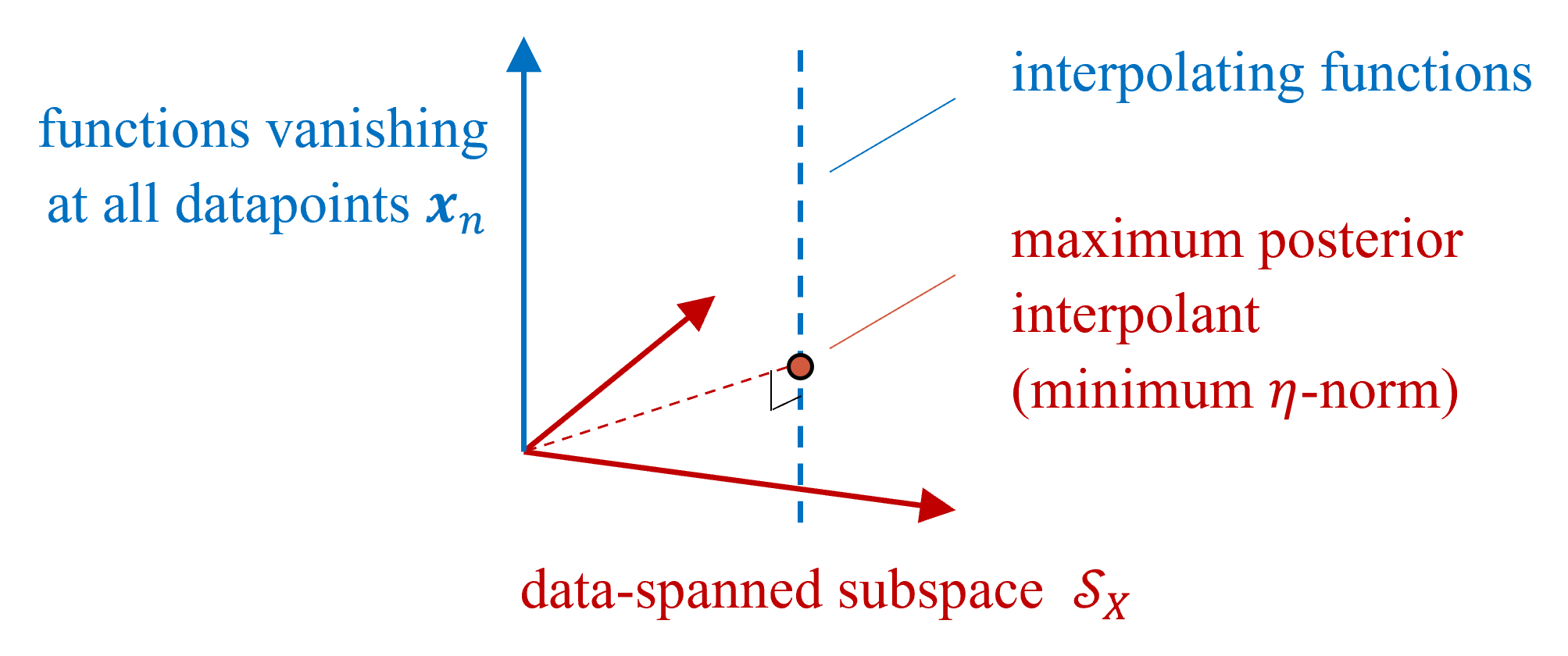}
	\caption{Subspaces of $\cF_\eta$} 
	\label{fig:data-span}
\end{figure}

In this section, we condition the scale invariant process on $N$ datapoints $((\bx_1,y_1), \dots, (\bx_N,y_N))=(X,\bm{y})$.
We first construct the subspace of $\cF_\eta$ that is spanned by the data (see Fig.\ \ref{fig:data-span}), formulate the maximum posterior interpolant, and finally consider the space of interpolants to derive the posterior process.

\subsection{Data-spanned subspace}

To construct the subspace $\cS_X\subset\cF_\eta$, 
we use the Green's functions $g_{\bxi}$ of the operator $O[\hat f]=\hat f(\bk)/a_\eta^2(\bk)$, which by definition satisfy $O[g_{\bxi}]=\delta_{\bxi}$.
The important property of these Green's functions is that they are $\eta$-biorthogonal to the delta functions in $\cF_\eta$
\begin{equation}
	\langle \delta_{\bxi_1},g_{\bxi_2}\rangle_\eta = \langle \delta_{\bxi_1},O[g_{\bxi_2}]\rangle = \delta_{\bxi_1 - \bxi_2} \label{biorthogonality} \\
\end{equation}
and hence the data-centered Green's functions $g_{\bx_n}$ are $\eta$-orthogonal to all interpolating functions.
The Green's functions read
\begin{align}
	\hat g_{\bxi}(\bk) &= a_\eta(\bk)\,\exp(2\pi i \,\bk\bxi) = \frac{\exp(2\pi i \,\bk\bxi)}{\|\bk\|^{2\eta+D}} \label{Greens_function} \\
	g_{\bxi}(\bx) &= \frac{1}{\tilde C} \|\bx-\bxi\|^{2\eta} \label{Greens_function_position}
\end{align}
where the position space representation (\ref{Greens_function_position}) follows from the Fourier pair of generalized functions established by \citet{gelfand_generalized_1964}
\begin{equation}
	\|\bx\|^\gamma \longleftrightarrow \frac{\tilde C}{\|\bk\|^{\gamma+D}} \qquad    
	\begin{array}{r}
		\gamma\in\R\phantom{,4,6,\dots} \\
		\gamma\neq2,4,6,\dots \\
		-(\gamma+D)\neq2,4,6,\dots
	\end{array} \label{Gelfand}
\end{equation}
The value of the constant $\tilde C$ is not important at this point.

One problem we have to deal with is that the Green's functions have infinite $\eta$-norm, i.e.\ they do not belong to $\cF_\eta$ themselves and can hence not be used directly to span $\cS_\eta$.
Fortunately however, it is possible to construct linear combinations of Green's functions that \emph{are} finite in $\cF_\eta$.
The result, which will be shown in the Appendix is that $\cS_X$ consists of all linear combinations of $g_{\bx_n}$ whose coefficients $\bb$ satisfy (\ref{GRC}).
\begin{align}
	f_X(\bx) &= \sum_{n=1}^{N} b_n g_{\bx_n}(\bx) \label{data-span} \\
	M\bb &= 0  \qquad\Longleftrightarrow\qquad |f_X|_\eta<\infty \label{GRC} \\
	M_{\bnu n} &= \bx_n^{\bnu} = \prod_{d=1}^D (\bx_n)_d^{\nu_d} \qquad|\bnu|<\eta \label{monomial_matrix} 
\end{align}
Note that the number of constraints (\ref{GRC}) equals the number of nullspace dimensions (\ref{N0}), and hence the dimension of $\cS_X$ is $N-N_0$.

To understand the situation intuitively, we point out that the infinite norm of $g_{\bxi}(\bx)$ can be attributed to the fact that they ``grow too fast'' as $\|\bx\|\rightarrow\infty$.
The linear combination (\ref{data-span}) ensures that the fast growing terms of different $g_{\bx_n}$ cancel out such that $f_X(\bx)$ is bounded by a multiple of $\|\bx\|^\eta$.

\subsection{Maximum posterior interpolation}
\label{sec:maxpost_interpolation}

For interpolation, we need to consider both $\cS_X$ and the nullspace $\mathcal{N}_\eta$.
Note that the nullspace and the constraints (\ref{GRC}) are dual, and the Cartesian product $\cS_X^* = \cS_X \times \mathcal{N}_\eta$ has exactly $N$ dimensions.

Altogether, we have $N+N_0$ degrees of freedom (coefficients of the Green's functions and the monomials), and $N+N_0$ constraints (interpolation and (\ref{GRC})).
Together, this yields the following linear system that describes the maximum posterior interpolation $f_{X,\by}$:
\begin{align}
	f_{X,\bm{y}}(\bx) = \sum_{n=1}^N & b_n g_{\bx_n}(\bx) + \sum_{|\bnu|<\lfloor\eta\rfloor} c_{\bnu}\bx^{\bnu} \label{interpolant} \\
	\begin{pmatrix} G & M^T \\ M & 0 \end{pmatrix}
	\begin{pmatrix} \bb \\ \bm{c} \end{pmatrix}
	&= \begin{pmatrix} \by \\ 0 \end{pmatrix} \label{ac} \\
	G_{nm} = g_{\bx_m}(\bx_n) &= \|\bx_n-\bx_m\|^{2\eta} \label{biorth_matrix}
\end{align}

\subsection{Polyharmonic splines} 

The result (\ref{interpolant}--\ref{biorth_matrix}) is known as polyharmonic spline interpolation in the field of few-dimensional data analysis \citep{wendland_scattered_2004}, \citep{fasshauer_meshfree_2007}).
There is an important difference concerning the range of valid $\eta$, however.
While polyharmonic splines are usually defined for positive integer and half-integer $\eta$ (positive integers of the typically used parameter $k=2\eta$), we find that the derivation holds for all positive (fractional) $\eta$ \emph{except} for the integers, and that the usually given solution for integer $\eta$ that involves $\|\bx\|^k\log\|\bx\|$ is incorrect.

The rather obvious hint that something is wrong is that the $\log\|\bx\|$ term introduces a distinguished scale although the problem statement is scale invariant.
Mathematically, the Green's functions are not unique for integer $\eta$, and the ``linear combination trick'' to obtain finite $\eta$-norm does not work anymore.
This issue will require a more thorough treatment elsewhere.

\subsection{Posterior process}
\label{sec:posterior_process}

To determine the shape of the posterior process, we consider the extended subspace $\cS_{XT}$ that is spanned by the $N$ datapoints and $N_T$ test points.
We will denote vectors in $\cS_X$ by $\bh$ and vectors in the orthogonal complement\footnote{Note that $\cS_T$ (spanned only by the test points) is \emph{not} $\eta$-orthogonal to $\cS_X$ and is therefore different from the orthogonal complement.}
$\cS_{XT}/\cS_X$ by $\bh_T$.
It is not required here to determine the components of $\bh$ and $\bh_T$, so we defer this to Sec.\ \ref{sec:Orthonormal_basis}.

The prior in $\cS_{XT}$ is hyperbolic-isotropic
\begin{align}
	p(\bh_{XT}) &= \frac{1}{\|\bh_{XT}\|^{N_h+N_T}} \label{TP_first}
\end{align}
and it is quite straightforward to see that its conditionals are Student $t$-distributions.
More concretely, let $\bh_{XT}=[\bh, \bh_T]$ be the complete vector in $\cS_{XT}$, then the conditional distribution $p(\bh_T|\bh)$ is a multivariate $t$-distribution with $N_h=N-N_0$ degrees of freedom (unless $\|\bh\|=0$):
\begin{align}
	p(\bh_T|\bh) &\propto \big( \|\bh_T\|^2 + \|\bh\|^2 \big)^{-\frac{N_h+N_T}{2}} \\
	&= \|\bh\|^{-N_{hT}} \left(1 + \frac{\|\bh_T\|^2}{\|\bh\|^2}\right)^{-\frac{N_h+N_T}{2}} \\
	&\propto p_\text{St}(\bh_T;\nu,\Sigma) \qquad
	\begin{array}{l}
		\nu = N_h \\
		\Sigma = \frac{\|\bh\|^2}{N_h} \, I_{N_T}
	\end{array} \label{TP_last}
\end{align}
Here, $p_\text{St}$ denotes the multivariate $t$-distribution, and $I_{N_T}$ is the identity matrix of size $N_T\times N_T$.
The constraint $\|\bh\|\neq0$ means that we need at least $N\geq N_0+1$ datapoints that cannot be interpolated with a nullspace polynomial.
Since the map between $\bh$ and $\by$ is linear, the function values $\by_T=f(\bx_T)$ also follow a multivariate $t$-distribution with $\nu=N_h$, albeit with a different $\Sigma$.
When taking the limit to infinitely many test points, the multivariate $t$-distribution becomes a $t$-process.






\subsection{Credible intervals} 
\label{sec:credible_intervals_interp}

\begin{figure}
	\centering
	\includegraphics[width=\linewidth]{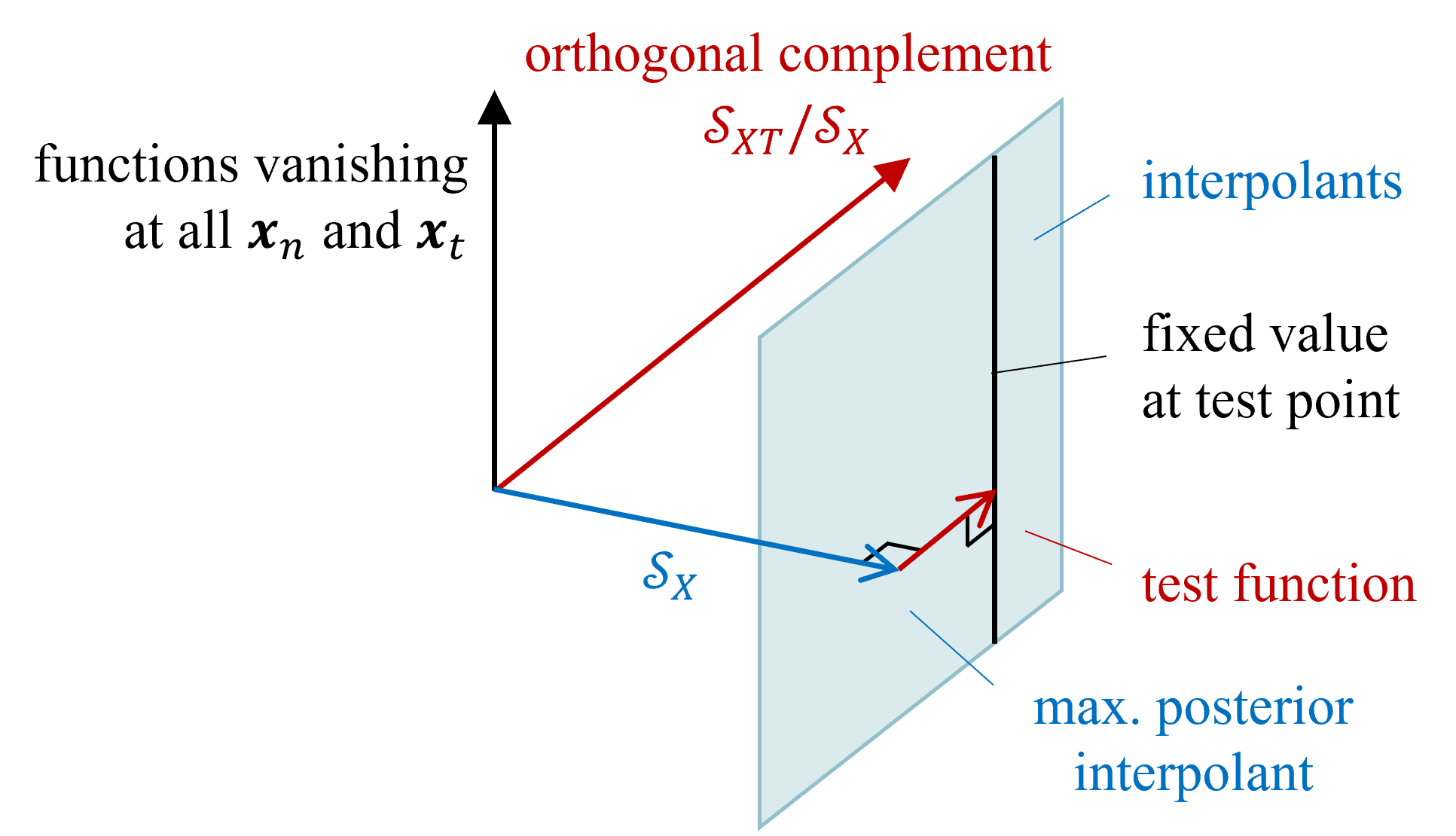}
	\caption{Test Function to Determine Credible Intervals}
	\label{fig:test_function}
\end{figure}

We now consider a single test point $\bx_t$ and determine the width of the posterior $t$-distribution of $y_t=f(\bx_t)$.
To do this, we need the scaling factor between $h_t$ (the single component of $\bh_T$) and $y_t$.
Fig.\ \ref{fig:test_function} illustrates the relation between the subspaces $\cS_X$ (blue) and $\cS_{XT}/\cS_X$ (red). 
The ``test function'' $f_t(\bx)$ that corresponds to $h_t$ can be characterized as having minimal $\eta$-norm among all functions that attain a certain value $y_t$ at the test point and vanish at all training points.
For simplicity we choose $y_t=f_{X,\by}(\bx_t)+1$, and the test function can now be determined in analogy to (\ref{interpolant}--\ref{biorth_matrix}) by:
%
\begin{align}
	f_t(\bx) =\; b_t g_{\bx_t}(\bx) + \sum_{n=1}^N b_n g_{\bx_n}(\bx&) + \sum_{|\bnu|<\lfloor\eta\rfloor}\!\!\!\! c_{\bnu}\bx^{\bnu} \label{test_function} \\
	\begin{pmatrix} 0 & \bm{g}^T & \bm{m}^T \\ \bm{g} & G & M^T \\ \bm{m} & M & 0 \end{pmatrix}
	\begin{pmatrix} b_t \\ \bb \\ \bm{c} \end{pmatrix}
	&=\begin{pmatrix} 1 \\ \bm{0} \\ \bm{0} \end{pmatrix} \\
	\bm{g} &= [g_{\bx_t}(\bx_n)]_n \label{test_biorth} \\
	\bm{m} &= [\bx_t^{\bnu}]_{\bnu}
\end{align}
The sought scaling factor between $h_t$ and $y_t$ is now given by $|f_t|_\eta$, which can be calculated by joining $b_t$ to $\bb$ and applying the following general formula for polyharmonic splines of the form (\ref{interpolant}):
\begin{align}
	|f_{\bb}|_\eta &= C\cdot \bb^T G \bb \label{eta-norm} \\
	C &= (-1)^{\lceil\eta\rceil} \frac{\Gamma\big(\eta+\frac{1}{2}\big) \;\pi^{\frac{D+1}{2}}} {\Gamma\big(\eta+\frac{D}{2}\big) \; \Gamma(2\eta+1)} \label{constant}
\end{align}
We prove this formula in the Appendix for positive half-integer $\eta$, and numerical evidence suggests that this result is indeed valid for all positive non-integer $\eta$.

Now we can write down the marginal posterior of $y_t$ as
\begin{align}
	p(y_t|\bx_t,X,\bm{y}) &= p_\text{St}\left(y_t;\nu,s(\bx_t)\right) \\
	\nu &= N_h = N - N_0 \\
	s(\bx_t) &= \frac{1}{\sqrt{\nu}} \frac{|f_{X,\bm{y}}|_\eta}{|f_t|_\eta} \label{t-scale} \\
	\sigma(\bx_t) &= \frac{1}{\sqrt{\nu-2}}\, \frac{|f_{X,\bm{y}}|_\eta}{|f_t|_\eta} \label{t-sigma}
\end{align}
where $s$ and $\sigma$ are the $t$-distribution's scale parameter and standard deviation, resp.

\section{Regression}
\label{sec:Regression}

In this section, we formulate the posterior process for a Gaussian likelihood. 
As with interpolation, we start in the subspace $\cS_X^*=\cS_X\times\mathcal{N}_\eta$, for which we first construct an orthonormal basis in order to formulate the posterior.
The orthogonal complement $\cF_\eta/\cS_X^*$ will be considered only in the last subsection.



\subsection{Orthonormal basis}
\label{sec:Orthonormal_basis}

Conceptually the simplest (though not the most efficient) approach to construct an $\eta$-orthonormal basis of $\cS_X$ 
is an iterative procedure using the test functions of Sec.\ \ref{sec:credible_intervals_interp}.
We start with a minimal set of $N_0\!+\!1$ datapoints $X_{N_0+1}=[\bx_1,\dots,\bx_{N_0+1}]$ that span a single-dimensional subspace $\cS_{X_{N_0+1}}$ consisting only of multiples of $f_{X_{N_0+1}}$, see (\ref{data-span}, \ref{GRC}).
Then we consecutively add datapoints $\bx_{n+1}$, each time constructing the corresponding ``test function'' (\ref{test_function}--\ref{test_biorth}), which is orthogonal to the previous space $\cS_{X_n}$.
We collect the coefficients $b_{mn}$ of $f_{X_{N_0+1}}$ and all test functions in the matrix $\tilde H$, leaving out the polynomial coefficients $\bm{c}$:
\begin{equation}
	\tilde H = \begin{pmatrix}
		b_{1,1} & \dots & b_{1,N_h-1} & b_{1,N_h} \\
		\vdots &  & \vdots & \vdots \\
		b_{N_0\!+\!1,1} &  & \vdots & \vdots \\
		0 & \ddots & \vdots & \vdots \\
		\vdots & \ddots & b_{N\!-\!1,N_h\!-\!1} & \vdots \\
		0 & \dots & 0 & b_{N,N_h} \\
	\end{pmatrix}
\end{equation}
After scaling all column vectors to unit $\eta$-norm by dividing by (\ref{eta-norm}), we obtain the $\eta$-orthonormal $H$.
\begin{align}
C\cdot H^T G H &= I_{N_h}
\end{align}

\subsection{Posterior in extended data-spanned space $\cS_X^*$}
\label{sec:extended_SX}

For the following we need the relation between the likelihood's mean $\bm{\mu}_y=\by$ and covariance $\Sigma_y$ in $\by$-space, and its mean $\bm{\mu}_h$ and covariance $\Sigma_h$ in $\cS_X^*$.  
To get an overview, we give a diagram of the two steps involved.
\begin{equation}
	\bh^* = \begin{bmatrix} \bh \\ \bm{c} \end{bmatrix}
	\text{\raisebox{.7em}{$\xrightarrow{H}$}}
	\begin{bmatrix} \bb \\ \bm{c} \end{bmatrix}
	\xrightarrow{[G,M^T]} \by
\end{equation}
The vectors $\bh^*$, $\bb$, and $\by$ have length $N$, $\bh$ has length $N_h$, and $\bm{c}$ has length $N_0$.
We will use the superscript star $^*$ as a general notation to indicate that the nullspace is included:
\begin{equation}
	\bb^* = \begin{bmatrix} \bb \\ \bm{c} \end{bmatrix}; \;\;\;
	H^* = \begin{bmatrix} H & 0 \\ 0 & I_{N_0} \end{bmatrix}; \;\;\;
	G^* = [G, M^T]
\end{equation}
Now we can directly write down:
\begin{align}
	E^* &= G^* H^* \label{E_matrix} \\
	\by_\mu &= E^* \bh_\mu^* \label{mu_y_h} \\
	\Sigma^{-1}_{h^*} &= (E^*)^T \Sigma^{-1}_y E^* \label{Sigma_y_h}
\end{align}
We obtain $\bh_\mu^*$ by solving (\ref{mu_y_h}) and $\Sigma_{h^*}^{-1}$ by inversion of $\Sigma_y$ and matrix multiplications.
This determines the core computational complexity of SIP regression, which is $\mathcal{O}(N^3)$ (or slightly lower if $\Sigma_y$ is diagonal), i.e.\ equal to GP regression.

The likelihood in $\cS_X^*$ now reads
\begin{align}
	p(\by|\bh^*,X) \propto \exp\!\left( \frac{1}{2} (\bh^*\!-\bh_\mu^*)^T\Sigma_{h^*}^{-1}(\bh^*\!-\bh_\mu^*) \!\right) \label{likelihood}
\end{align}
The prior over $\bh$ is hyperbolic-isotropic (as stated before).
Over the polynomial coefficients $\bm{c}$, we choose a uniform prior, which is justified by translation invariance of $y$ and the linearity of the polynomials w.r.t.\ their coefficients.
Together we have
\begin{align}
	p(\bh^*) \propto \frac{1}{\|\bh\|^{N_h}} \label{prior}
\end{align}
which has a cylindrical symmetry (note the difference between $\bh$ and $\bh^*$).
Combining (\ref{prior}) and (\ref{likelihood}), we obtain the log posterior (with some constant $C^*$)
\begin{align}
	\log p(\bh^*|X,\by) =\; &C^* - N_h \log\|\bh\| \dots \nonumber \\
	&- \frac{1}{2} (\bh^*-\bh^*_\mu)^T \Sigma_{h^*}^{-1} (\bh^*-\bh^*_\mu) \label{logpost}
\end{align}


\subsection{Posterior mean and covariance}
\label{sec:sampling}

The maximum posterior can be calculated quite easily, but we found experimentally that it is far off from the mean posterior and hence not a good estimate.
The reason is that the hyperbolic prior in high dimensions grows very strongly towards the origin, and hence the posterior is highly asymmetric.
Therefore, we need to find the mean posterior, which is an $N$-dimensional integral over $\cS_X^*$.
This can be done with a sampling method (for details see Sec.\ \ref{sec:Results}), from which we obtain estimates of the posterior mean $\hat\bh^*$ and covariance $\hat\Sigma_{\bh^*}$:

\subsection{Nullspace pole}
\label{sec:nullspace_pole}

\begin{figure}
	\centering
	\includegraphics[width=.8\linewidth]{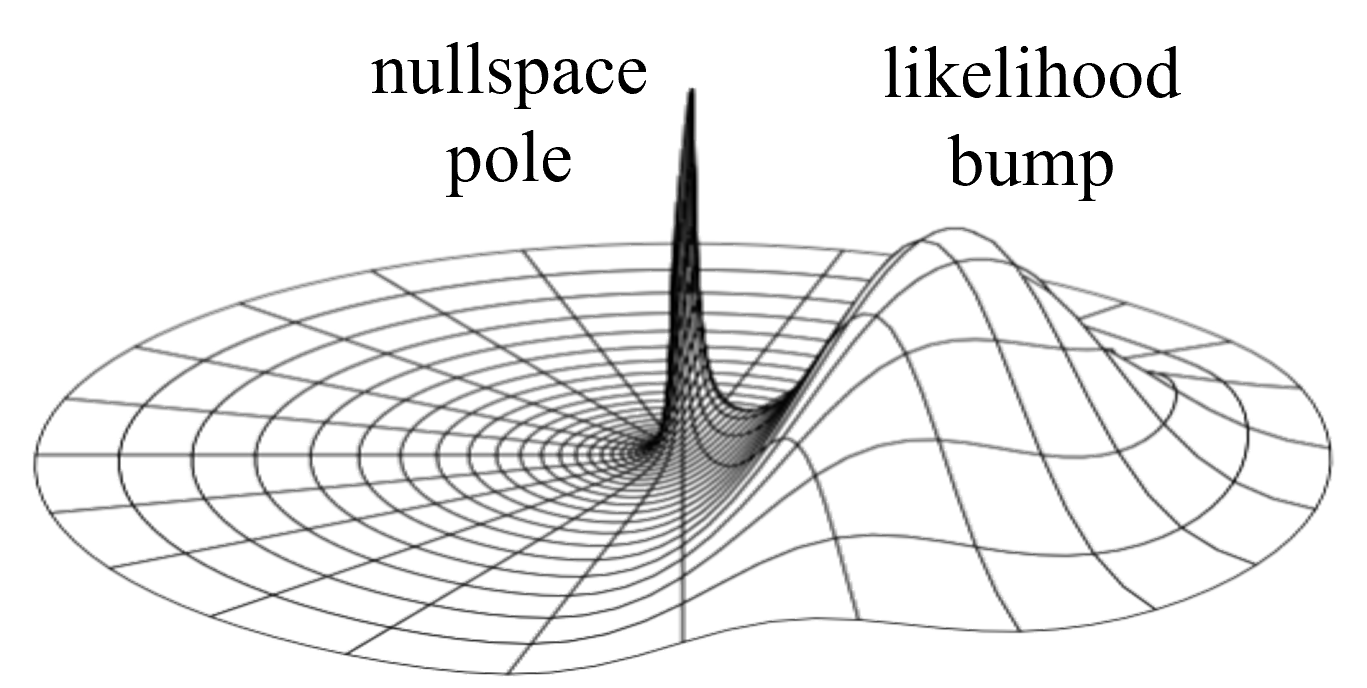}
	\caption{Sketch of Posterior in $\cS_X$.}
	\label{fig:nullspace_pole}
\end{figure}

It is important to note that the prior's pole at $\|\bh\|=0$ does not get canceled by the Gaussian likelihood and retains its infinite probability mass.
Strictly speaking, this means that SIP regression is reduced to polynomial regression within the nullspace.
For typical datasets, however, the ``nullspace pole'' and ``likelihood bump'' are well separated (see Fig.\ \ref{fig:nullspace_pole}), and the sampling procedure never encounters the nullspace pole.
A more complete treatment of this problem will require additional work.


\subsection{Unknown observation error}

The size of the observation error is typically not known in applications, which we model as
\begin{align}
	\Sigma_y &= \sigma_y^2 I_N & p(\sigma_y) &\propto \frac{1}{\sigma_y} \label{sigma_prior}
\end{align}
where $\sigma_y$ is a hyperparameter that we sample together with $\bh^*$.
The distribution of $\bh^*$ depends on $\sigma_y$, but that does not pose a problem for the sampling procedure.

Note that the prior over $\sigma_y$ has the same issue as the prior over $\bh^*$:
The pole does not get canceled during inference, which means that strictly speaking the posterior assigns infinite probability mass to interpolation.
As before however, this pole is usually well-separated from the ``likelihood bump'' so that the sampling procedure does not encounter it.

\subsection{Posterior process and credible intervals}
\label{sec:credible_intervals}

To obtain the complete posterior process, we now include the orthogonal complement $\cF_\eta/\cS_X^*$ into our consideration.
By combining the posterior in $\cS_X^*$ with the $t$-process in $\cF_\eta/\cS_X^*$, we obtain an $N$-dimensional mixture of $t$-processes with $N_h$ degrees of freedom and different means $f_{\bh^*}\!(\bx)$ and scale factors $s_{\bh^*}\!(\bx)$
\begin{align}
	p(f|X,\by) &= \!\int\!\! p_\text{St}\big(f; N_h, f_{\bh^*}\!, s_{\bh^*}\!\big) \; p(\bh^*\!|X,\by) \dd\bh^* \label{regression_process} \\
	s_{\bh^*}(\bx_t) &= \frac{1}{\sqrt{\nu}} \frac{|f_{\bh^*}|_\eta}{|f_t|_\eta} = \frac{\sqrt{\nu-2}}{\sqrt{\nu}}\, \sigma_{\bh^*}(\bx_t)
\end{align}
For our current purpose it is sufficient to approximate (\ref{regression_process}) with a single $t$-process centered at the mean posterior $f_{\hat\bh^*}$.
As approximate variance $\sigma_f^2$ we use the squared sum of two terms $\sigma_t^2$ and $\sigma_s^2$ that represent the contributions of $\cF_\eta/\cS_X^*$ and $\cS_X^*$, resp.
\begin{align}
\sigma_f^2(\bx) &= \sigma_t^2(\bx) + \sigma_s^2(\bx) \\
\sigma_t(\bx_t) &= \sigma_{\hat\bh^*}(\bx_t) \\
\frac{1}{\sigma_s^2(\bx_t)} &= \bx_t^T (E^*)^{-T} \,\hat\Sigma_{\bh^*}^{-1}\, (E^*)^{-1}\, \bx_t
\end{align}
$\sigma_t^2$ is the variance of the $t$-process at the mean posterior $\hat\bh^*$, and $\sigma_s^2$ is the result of transforming the estimated covariance $\hat\Sigma_{\bh^*}$ from $\cS_X^*$ back to $\by$-space.
In the following we call $\sigma_s$ the \emph{spline uncertainty}, $\sigma_t$ the \emph{interpolation uncertainty}, and $\sigma_f$ the \emph{function uncertainty}.

Note that $\sigma_f$ describes the uncertainty of our estimate of the true function $f(\bx)$ and does \emph{not} contain the data noise $\sigma_y$.
For some purposes it is useful to add the two terms to obtain the \emph{data uncertainty} $\sigma_d$, which is also the quantity typically reported in GP regression.
\begin{equation}
	\sigma_d^2(\bx) = \sigma_f^2(\bx) + \sigma_y^2(\bx)
\end{equation}

\section{Results}
\label{sec:Results}

\paragraph{Implementation details}

The following results have been obtained by sampling with the NUTS 
sampler of pymc3 \citep{salvatier_probabilistic_2016}.
We use 2 chains with 1000 samples each, of which 500 are discarded as burn-in.
As initial values we use $\bh_{\text{init}}^*=\bh_\mu^*$ and $\sigma_{y,\text{init}} = \sqrt{\text{Var}(\by)} / 10$.
To improve sampling efficiency, we reparametrize $\bh^*$ with the Cholesky decomposition of $\Sigma_{h^*}^{-1}$, which reduces the correlations between vector components.

Since automatic relevance determination (ARD) is beyond the scope of this paper, we scale all features to equal range (min-max).
For datasets with discrete features, we add a small amount of jitter (0.1\% of the range) to avoid numerical problems.


For the comparison with GP regression, we use \citet{gpy_gpy_2012}, always optimizing the kernel parameters with the default settings.

\paragraph{Performance}

\begin{table*}
	\centering
	\caption{Performance Comparison on Different Datasets: RMSE of 5-fold Cross-Validation.}
	\bigskip
	\begin{small}
		\begin{tabular}{ll|ccccccc}
			& & marathon & diabetes & boston & energy eff. 0 & energy eff. 1 & concrete & red wine \\ \hline
			& \# dims & 1 & 10 & 13 & 8 & 8 & 8 & 11 \\
			& \# datapoints & 27 & 442 & 506 & 768 & 768 & 1030 & 1599 \\ \hline
			GP & Matérn 1/2 & 0.243 & \textbf{54.32} & 2.92 & 1.86 & 2.22 & 5.83 & \textbf{0.58} \\
			GP & Matérn 3/2 & \textbf{0.219} & 54.33 & \textbf{2.83} & 0.75 & 1.36 & \textbf{5.65} & 0.61 \\
			GP & Matérn 5/2 & 0.235 & 54.36 & 2.88 & 0.68 & 1.43 & 5.74 & 0.62 \\
			GP & rational quad. & 0.234 & 64.91 & 2.91 & \textbf{0.54} & \textbf{1.32} & 5.83 & 0.58 \\
			GP & squared exp. &  0.238 & 54.37 & 3.00 & 0.73 & 1.50 & 5.83 & 0.62 \\ \hline
			SIP & $\eta=1/2$ & 0.235 & \textbf{54.23} & 2.97 & 1.95 & 2.31 & 5.83 & \textbf{0.58} \\
			SIP & $\eta=1.01$ & \textbf{0.215} & 54.60 & \textbf{2.86} & 1.10 & 1.62 & 5.65 & 0.63 \\
			SIP & $\eta=3/2$ & 0.227 & 54.56 & 2.98 & 0.77 & \textbf{1.36} & \textbf{5.62} & 0.71 \\
			SIP & $\eta=2.01$ & 0.556 & 54.50 & 3.38 & 0.70 & 1.51 & 5.70 & 0.64 \\
			SIP & $\eta=5/2$ & 0.861 & --- & 3.69 & \textbf{0.62} & 1.69 & 30.49 & 1.22
		\end{tabular}
	\end{small}
	\label{tab:RMSE}
\end{table*}

Table \ref{tab:RMSE} shows the prediction error of the posterior mean on several well-known benchmark datasets.
The performance of the best variants of SIP and GP are very similar for all datasets.
Also, there is a nice agreement concerning which kernel/$\eta$ is best for a given dataset; GP and SIP give similar ``estimates'' of the regularity of the true function.


\paragraph{Differences to GP regression}
\label{sec:tanh}

\begin{figure}
	\centering
	\includegraphics[width=\linewidth]{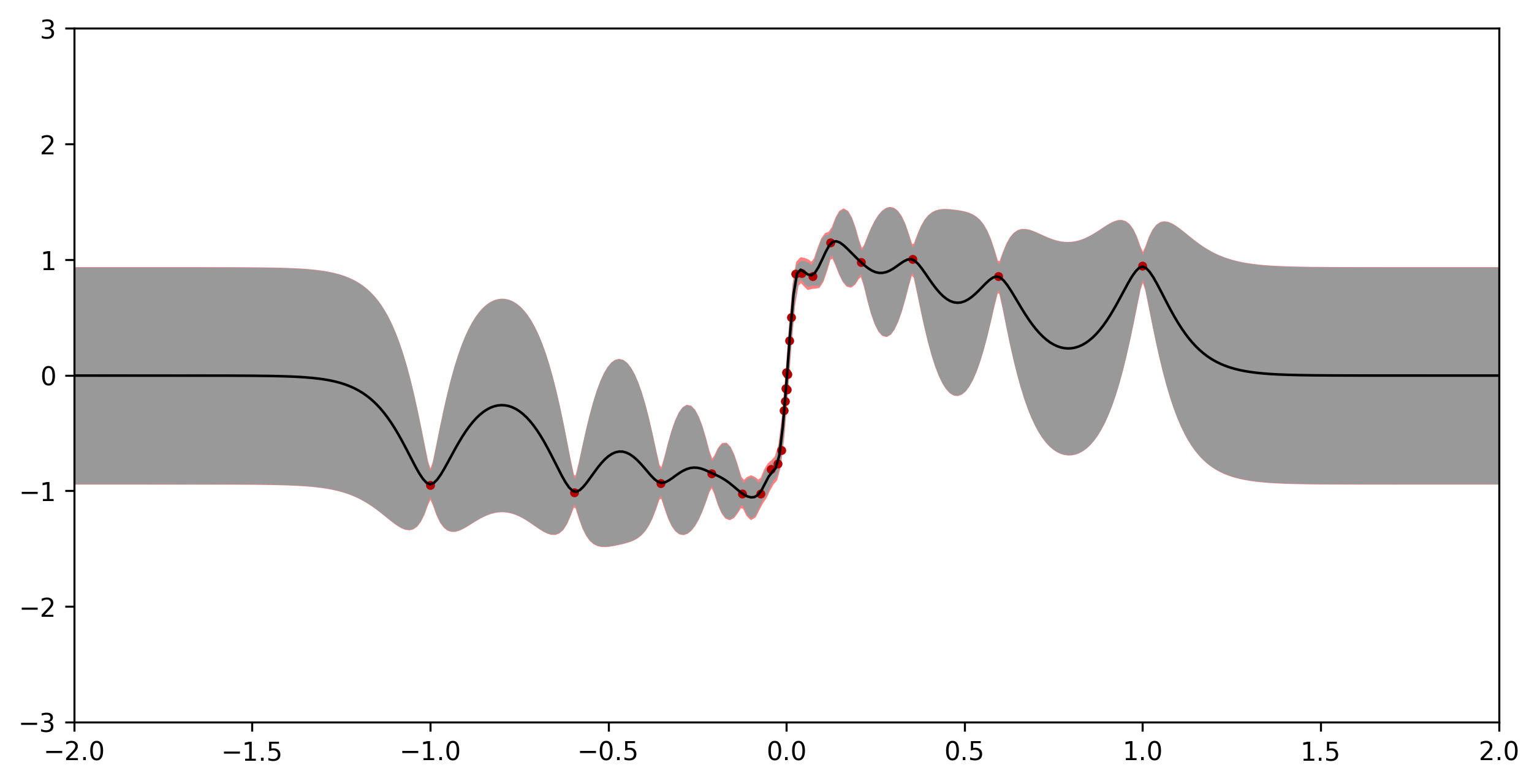}
	\includegraphics[width=\linewidth]{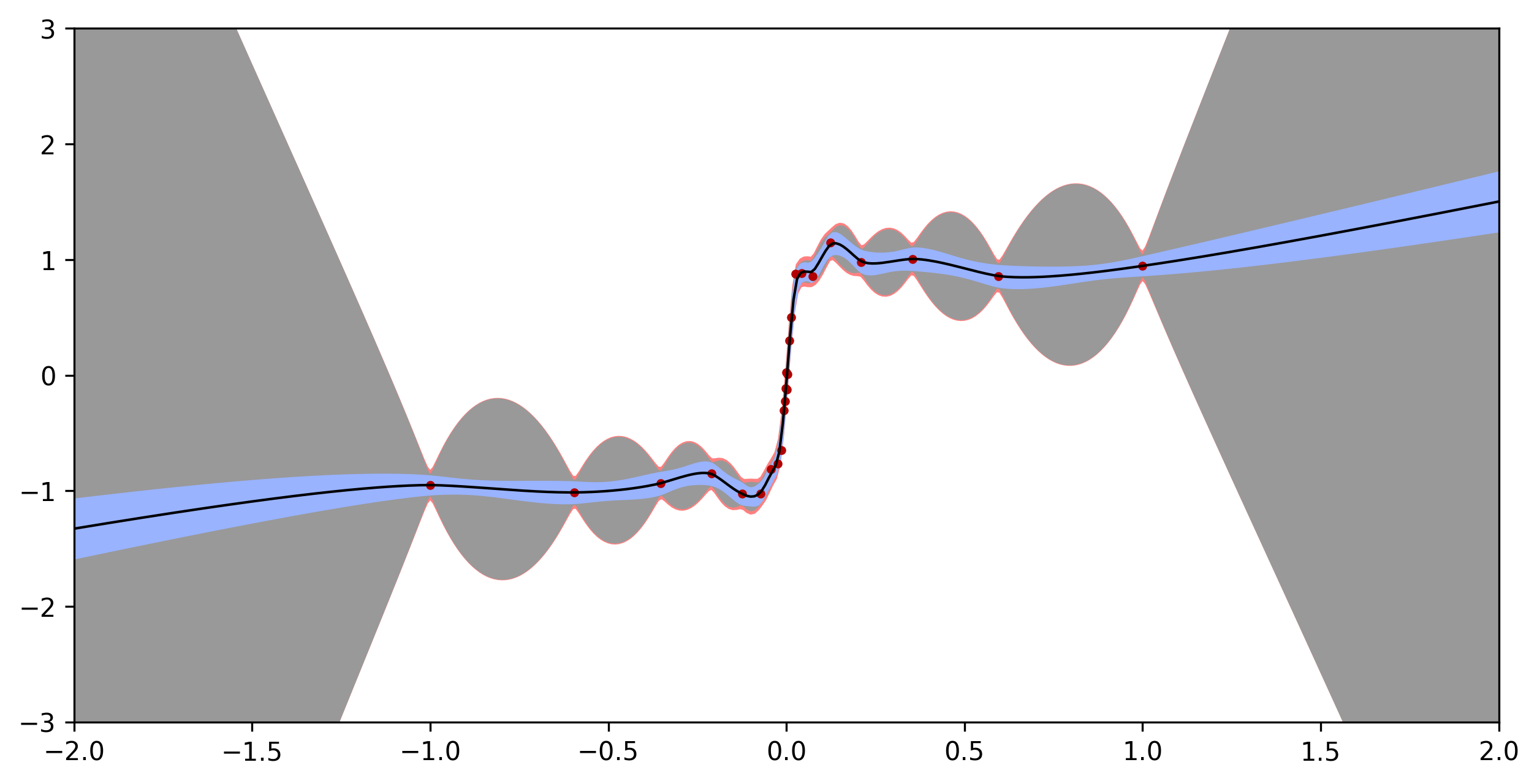}
	\caption{Comparison on Non-Stationary Data. Top: GP, rational quadratic. Bottom: SIP, $\eta=1.01$}
	\label{fig:tanh}
\end{figure}

In most cases, the results of GP and SIP are very similar.
To highlight the differences, we use a synthetic step dataset in Fig.\ \ref{fig:tanh}, although neither SIP nor vanilla GP are really suitable for this kind of non-stationary data.

The sharp step forces both methods to learn a very small length scale, which leads to overfitting in regions of scarce data.
This problem is more severe for GPs, however, because their posterior mean is biased towards the prior mean, which is obviously inappropriate in this case. 
We want to point out that the term ``prior'' is somewhat inappropriate in this context, because the ``prior'' mean is in fact optimized to the data.

The different types of credible intervals are shown as blue (spline uncertainty), gray (function uncertainty), and red (data uncertainty; hardly visible because almost equal to the function uncertainty in this case).
For extrapolation outside the data range, the SIP credible intervals increase quickly, which correctly represents the lack of information in this region, while the credible intervals of stationary GPs approach the constant learned variance, which is obviously inappropriate. 
We believe that the extrapolation behavior of SIP will prove beneficial for applications such as Bayesian optimization.




\section{Discussion}

\paragraph{Regularization parameters vs.\ hyperparameters}

For the following discussion, we distinguish between regularization parameters that \emph{must} be adapted to each dataset in order to obtain a reasonable result, and other ``proper'' hyperparameters that can be set to a default value when the costly optimization is not absolutely necessary.
For GP regression, the kernel parameters (length scale, variance, prior mean) are regularization parameters, while the kernel shape is a ``proper'' hyperparameter.
For SIP regression, there is no regularization parameter at all and only one hyperparameter $\eta$.

\paragraph{Regularization determined by scale invariance}

We want to point out that it is quite remarkable that SIP regression works at all without any regularization parameter.
In the log posterior (\ref{logpost}), the role that is usually played in comparable models by a regularization parameter $\lambda$ is now fulfilled by $N_h$, which is a fixed value derived from the principle of scale invariance.
The fact that $N_h=N-N_0$ leads to a good tradeoff between data fit and regularization on all the datasets we tried is the most significant result of this paper, regardless of whether the RMSE turns out a little bit better or worse than GP regression.

Of course, regularization is also affected by the hyperparameter $\eta$, but only in an indirect way and not very strongly.
We typically get decent results for a standard value of, say, $\eta=3/2$.

\paragraph{Practical advantages}


The fact that SIP regression has no regularization parameter is also a significant practical advantage.
For GP regression, the optimization of the kernel parameters takes up the main share of the computational cost, because it requires fitting the GP model many times. 
For SIP regression, no such optimization is required.

The baseline sampling procedure used in this paper is quite slow and does not yet realize this potential, but the simple form of the posterior (\ref{logpost}) allows a more efficient inference procedure, which will be the topic of a subsequent publication.

Concerning the hyperparameters, we point out that the space of possible kernels for GP regression is huge and cannot be explored completely, for SIP regression on the contrary, the single hyperparameter $\eta$ can be optimized quite easily if required.
One might argue that the large space of possible GP kernels can also be an advantage, because it allows to tailor the GP to specific applications.
While this is true, we want to point out that for the majority of applications there is no clear indication to guide the choice of GP kernel, and in most cases one takes one of the standard kernels simply because one \emph{must} make \emph{some} choice in order to apply GPs.
We would actually argue that the forced choice of kernel is a disadvantage, because it introduces a certain arbitrariness into the method.
The choice of kernel, and hence the result will depend to a certain extent on the ML practitioner's personal experience, taste, and convenience.
We believe that in the case of missing prior information about a suitable kernel it is better to use SIP, which is invariant and hence ``uninformative'' and ``non-prejudiced''.


\paragraph{Outlook}

SIP regression assumes Euclidean structure of input and output space, which is not always appropriate.
Especially, some form of automatic relevance determination (ARD) is often required instead of assuming rotation invariant (isotropic) input space.
Another issue is translation invariance of the input, which corresponds to stationarity.
For many applications it is beneficial to allow some form of non-stationarity.
Future work will have to address these two issues.

Lastly, we mention that the idea of using a scale invariant prior can probably be extended to other ML areas beyond regression, like classification, density estimation, or clustering.

\bibliography{references}  

%
%
%

\onecolumn
\aistatstitle{Supplementary material for:\\ Scale invariant process regression: Towards Bayesian ML with minimal assumptions}

\section{ADDITIONAL RESULTS}

\subsection{Fits on marathon dataset}

\begin{figure}[h]
	\centering
	\includegraphics[width=.49\textwidth]{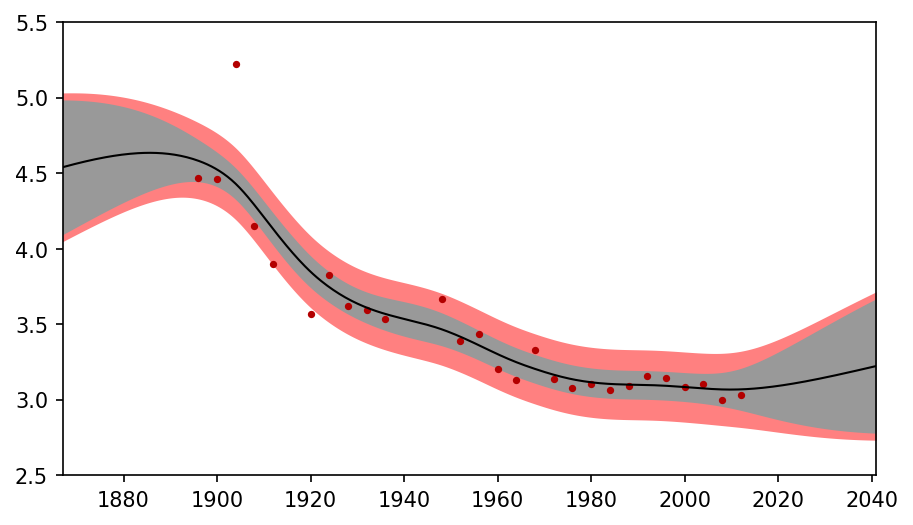}
	\includegraphics[width=.49\textwidth]{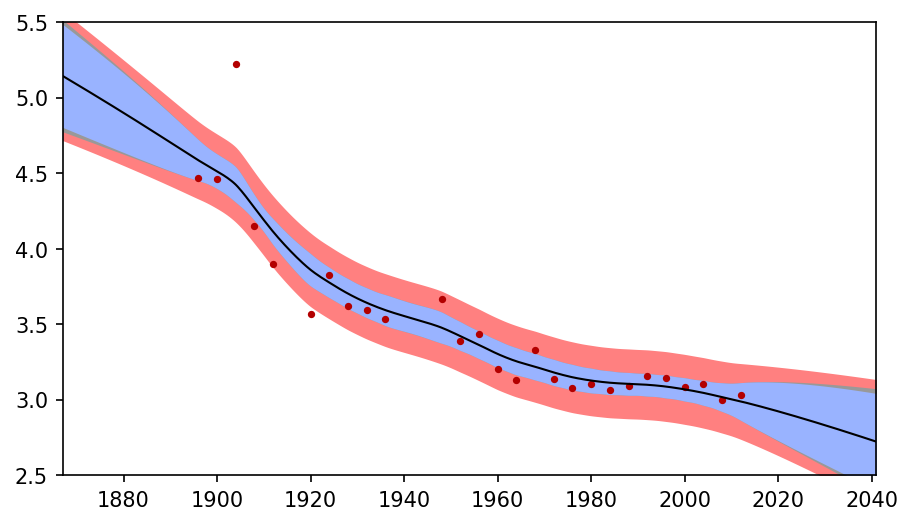}
	\caption{Best Fits on Marathon Dataset. Left: GP, Matérn 3/2; right: SIP $\eta=1.01$.}
	\label{fig:marathon}
\end{figure}

Figure \ref{fig:marathon} shows fits of the best GP model and the best SIP model on the well-known Marathon dataset.
The corresponding RMSEs of 5-fold cross-validation have been reported in Table 1 of the main paper.
Here, we show fits on the complete dataset, since this is the best estimate we can make in practice.

Note that for SIP, the function uncertainty is dominated by the spline uncertainty (blue), which is in contrast to Figure 4 of the main paper, where the function uncertainty was dominated by the interpolation uncertainty (gray).
Which type of uncertainty dominates depends on the (prescribed or estimated) observation error $\sigma_y$ and on how evenly spaced the datapoints are.

Besides the extrapolation behavior which has been discussed in the main paper, the most noticeable difference between the two models shown is that the outlier at 1904 causes a noticeable kink in the SIP fit, while it does not seem to affect the GP fit at all.

\begin{wrapfigure}{r}{6.5cm}
	\centering
	\vspace{-1.2cm}
	\includegraphics[width=.35\textwidth]{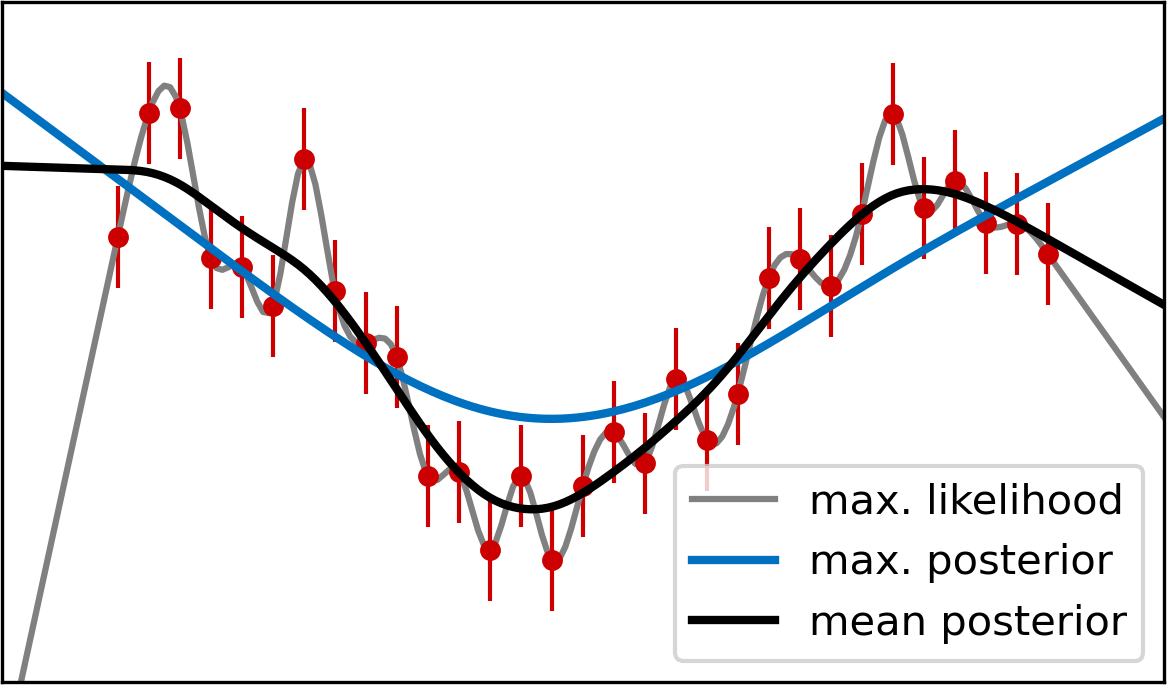}
	\vspace{-.2cm}
	\caption{Comparison between max.\ posterior and mean posterior.}
	\vspace{-2.5cm}
	\label{fig:maxpost}
\end{wrapfigure}

\subsection{Maximum posterior vs. mean posterior}

As stated in Section 4.3, the maximum posterior of SIP is not a good estimate, because the posterior is highly asymmetric.
The maximum posterior typically has much smaller $\eta$-norm than the mean posterior and shows significant overregularization.
An example of this behavior is shown in Figure \ref{fig:maxpost} on a sandbox dataset.

\vspace{2cm}
\pagebreak

\begin{wrapfigure}{r}{8.5cm}
	\centering
	\vspace{-1.5cm}
	\includegraphics[width=.49\textwidth]{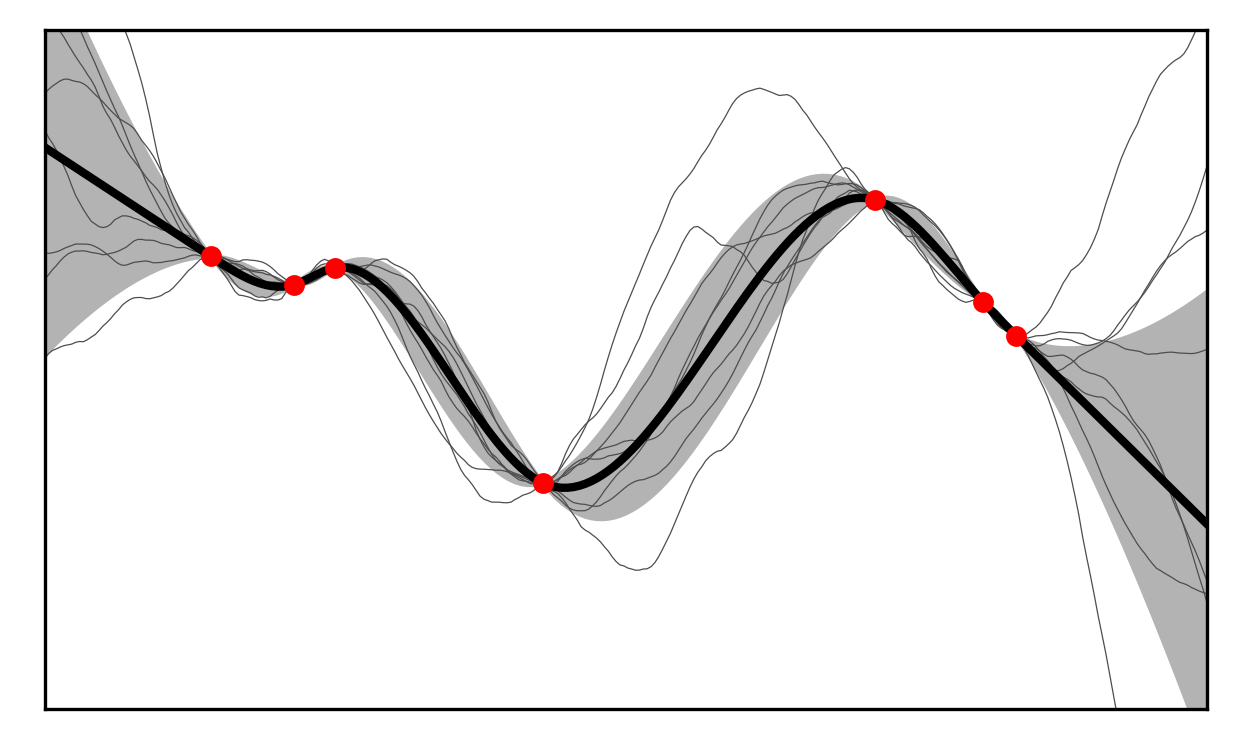}
	\vspace{-.5cm}
	\caption{Sample paths of the posterior process.}
	\vspace{-1.5cm}
	\label{fig:sample_paths}
\end{wrapfigure}

\subsection{Sample paths}

For completeness, we show some exemplary sample paths of the posterior process in Figure \ref{fig:sample_paths}.
The paths have been calculated by the expensive but simple procedure of alternately (i) drawing from the pointwise posterior ($t$-distribution) and (ii) recalculating the pointwise posterior including the new datapoint.

The regularity parameter was chosen as $\eta=3/2$, and the sample paths are quite similar to sample paths of a GP with a Mat\'ern 3/2 kernel.

\vspace{1cm}

\section{ADDITIONAL DERIVATIONS and PROOFS}

In the main paper, we have stated several results, for the derivation of which we have referred to the Appendix.
These derivations/proofs are given in the following.

\subsection{Marginal consistency of scale invariant distributions}
\label{app:SI_marginals}

In this section, we show that all marginals of a scale invariant distribution are themselves scale invariant.
This fact follows from a simple scaling argument.

Concretely, let $p(\bx)$ be the improper scale invariant distribution in $\R^N$, then all marginal distributions over a subset of dimensions $\nu\in\{1,\dots,N\}$ parametrized by $\bx_\nu=[x_i:{i\in\nu}]$ are also scale invariant.
\begin{align}
	\left.\begin{aligned}
		p(\bx) &= \frac{1}{\|\bx\|_a^N} \\
		\|\bx\|_a^2 &= \sum_{i=1}^N \left(\frac{x_i}{a_i}\right)^2
	\end{aligned}\quad\right\rbrace
	\qquad \xRightarrow{\forall\nu\subset\{1,\dots,N\}} \qquad
	\left\lbrace\quad\begin{aligned}
		p(\bx_\nu) &= \frac{1}{\|\bx_\nu\|_a^{|\nu|}} \\
		\|\bx_\nu\|_a^2 &= \sum_{i\in\nu} \left(\frac{x_i}{a_i}\right)^2
	\end{aligned}\right.
\end{align}

For the proof, let $U$ be an $(N-|\nu|)$-dimensional subset of $\R^N$, which means that when integrating over $U$, the volume element $\dd^N\bx$ scales as
\begin{align}
	\bx &= s\bxi \\
	\int_U \dd^N\!\bx &= \int_U s^{N-|\nu|} \dd^N\!\bxi \label{dV}
\end{align}
Now we compare the integrals of $p(\bx)$ over $U$ and over the scaled subset $sU= \{s\bx:\bx\in U\}$:
\begin{align}
	I_U &= \int_U \frac{1}{\|\bx\|_a^N} \dd^N\bx \\
	I_{sU} &= \int_{sU} \frac{1}{\|\bx\|_a^N} \dd^N\bx = \int_U \frac{1}{s^N\|\bxi\|_a^N}\; s^{N-|\nu|}\dd^N\bxi = \frac{1}{s^{|\nu|}} I_U \label{SI}
\end{align}
To obtain the proposition, we only have to choose $U$ as the axes-parallel hyperplane that intersects $\bx_\nu$ and is parallel to all axes with index not in $\nu$.


\subsection{Regularity of scale invariant process}
\label{app:Regularity}

In this section, we show that the non-degenerate sample paths $f(\bx)$ of the $D$-dimensional scale invariant process with amplitude spectrum $a_\eta(\bk)\propto |\bk\|^{-(\eta+D/2)}$ have regularity $\eta$ almost surely.

The ``degenerate'' sample paths excluded above are the nullspace polynomials (which have zero $\eta$-norm) and all non-locally-bounded sample paths (which have infinite frequency amplitudes).
Throughout our work, the term ``regularity $\eta$'' is short for: $\lfloor\eta\rfloor$ times differentiable, with the $\lfloor\eta\rfloor$-th derivative being locally $\alpha$-H\"older continuous for all exponents $\alpha<\eta-\lfloor\eta\rfloor$.

In the following, we consider the sample paths over a bounded domain, for convenience chosen as the hypercube ${\|\bx-\bxi\|_\infty \leq 1}$, where $\bx$ is the fixed center and $\bxi$ the free variable.
The corresponding domain in frequency space is the integer grid $\bk\in\mathbb{Z}^D$, and the $\eta$-norm of the non-degenerate sample paths on the hypercube is finite almost surely.
\begin{align}
	|f|_{\eta,\bx} = \sum_{\bk\in\mathbb{Z}^D} \left| \frac{f(\bk)}{a_\eta(\bk)} \right|^2 < \infty
\end{align}
Note that $f(\bx)$ is real and $f(\bk)$ is complex with symmetry $f(\bk)=f^*(-\bk)$.

The following proof uses two equivalent representations of the fractional Laplacian (Kwa\'{s}nicki, 2017): 
as an integral in position space and as a Fourier transform.
The Laplace exponent $m$ must be positive and neither an integer nor a half-integer.
\begin{align}
	(-\Delta)^m f(\bx) &\propto \int_{\|\bx-\bxi\|_\infty<1} \frac{f(\bx)-f(\bxi)}{\|\bx-\bxi\|^{2m+D}} \dd\bxi \label{fracLaplace_integral} \\
	&= \sum_{\bk\in\mathbb{Z}^D} \|\bk\|^{2m} \,\exp(i2\pi\bk\bx) \,f(\bk) \label{fracLaplace_Fourier}
\end{align}
We will relate the singular integral representation (\ref{fracLaplace_integral}) to the H\"older condition, and the Fourier representation (\ref{fracLaplace_Fourier}) to the spectrum, which together implies the equivalence between spectrum $a_\eta$ and regularity $\eta$.

\paragraph{Local H\"older continuity and fractional Laplacian}

The regularity $\eta$ (differentiability pus H\"older continuity of the derivative) can be described as:
\begin{align}
	|f(\bx)-f(\bxi)| \leq C \|\bx-\bxi\|^\eta \qquad\text{for } \|\bx-\bxi\|_\infty<1
\end{align}
Inserting this into (\ref{fracLaplace_integral}), we obtain
\begin{align}
	(-\Delta)^m f(\bx) &\propto \int_{\|\bx-\bxi\|_\infty<1} \frac{f(\bx)-f(\bxi)}{\|\bx-\bxi\|^{2m+D}} \dd\bxi \\
	&\leq \int_{\|\bx-\bxi\|_\infty<1} \frac{|f(\bx)-f(\bxi)|}{\|\bx-\bxi\|^{2m+D}} \dd\bxi \\
	&\leq C \int_{\|\bx-\bxi\|_\infty<1} \frac{1}{\|\bx-\bxi\|^{2m+D-\eta}} \dd\bxi \\
	&\leq \infty \qquad\text{iff } 2m<\eta \label{Holder_continuity}
\end{align}
i.e.\ the fractional Laplacian of order $m$ is finite iff $f(\bx)$ is $\eta$-H\"older continuous with $\eta>2m$.

\paragraph{Fractional Laplacian and frequency spectrum}

The Fourier representation of the fractional Laplacian can be written as an inner product in the scale invariant process's function space $\cF_\eta$.
\begin{align}
	(-\Delta)^m f(\bx) &= \sum_{\bk\in\mathbb{Z}^D} \|\bk\|^{2m} \,\exp(i2\pi\bk\bx) f(\bk) \\
	&= \sum_{\bk\in\mathbb{Z}^D} \frac{\|\bk\|^{2m-2\eta-D}}{\|\bk\|^{2\eta+D}} \,\exp(i2\pi\bk\bx) f(\bk) \dd\bk \\
	&= \Big< \|\bk\|^{2m-2\eta-D}\exp(i2\pi\bk\bx), f(\bk) \Big>_\eta \label{frac_Laplacian}
\end{align}

As mentioned above, $f$ has finite $\eta$-norm almost surely, and therefore (\ref{frac_Laplacian}) will be finite almost surely iff the functions $L_{\bx}=\|\bk\|^{2m-2\eta-D} \exp(i2\pi\bk\bx)$ have finite $\eta$-norm.
This is the case iff
\begin{align}
	\Big| L_{\bx} \Big|_\eta^2 = \Big| \|\bk\|^{2m-2\eta-D} \,\exp(i2\pi\bk\bx) \Big|_\eta^2 &< \infty \\
	\sum_{\bk\in\mathbb{Z}^D} \frac{\|\bk\|^{2(2m-2\eta-D)}} {\|\bk\|^{-(2\eta+D)}} \dd\bk &< \infty \label{L-integral} \\
	4m-2\eta-D &< -D \\
	2m &< \eta
\end{align}
Since this is the same condition as in (\ref{Holder_continuity}), we can conclude the equivalence between spectrum $a_\eta$ and regularity $\eta$.

\subsection{Nullspace polynomials}
\label{app:Nullpoly}

In this section, we show that polynomials of degree strictly smaller than $\eta$ have zero $\eta$-norm.
We perform the calculation in frequency space, where the monomials are represented as derivatives of the delta-function $\delta(\bk)$.
With $\bx\in\R^D$ and a multi-index $\bnu=[\nu_1,\dots,\nu_D]$, we can write
\begin{align}
	f_{\bnu}(\bx) &= \bx^{\bnu} = \prod_{d=1}^D x_d^{\nu_d} \\
	f_{\bnu}(\bk) &= \bm{\partial}^{\bnu}\delta(\bk) = \prod_{d=1}^D \partial_d^{\nu_d} \delta(k_d)
\end{align}
The central idea of the proof is that the SI process's amplitude spectrum $a_\eta(\bk)=\|\bk\|^{-(\eta+D/2)}$ has a zero at $\bk=0$ which cancels delta-function derivatives up to a certain order.

For the calculation we treat the $\delta$-derivatives as the limit of a Gaussian function $g(\bk)$ with infinitesimal width.
For the scaling argument below it is convenient to separate the normalization factor $s^D$ from the function shape $g(\bk)$.
\begin{align}
	g(\bk) &= \frac{1}{(2\pi)^{D/2}}\exp\left(-\frac{1}{2}\|\bk\|^2\right) \\
	\delta(\bk) &= \lim_{s\rightarrow\infty} s^D \,g(s\bk) \\
	\bm{\partial}^{\bnu}\delta(\bk) &= \lim_{s\rightarrow\infty} s^D \,s^{|\bnu|} \,g^{(\bnu)}(s\bk)
\end{align}
where the factor of $s^{|\bnu|}$ with $|\bnu|=\sum_d\nu_d$ comes from the chain rule of differentiation.

Now we can calculate the $\eta$-norm of $f_{\bnu}(\bk)$ as follows, where we substitute $\bkk=s\bk$ in (\ref{substitution}):
\begin{align}
	|f_{\bnu}|_\eta^2 &= \int_{\R^D} \left(\frac{f_{\bnu}(\bk)}{a_\eta(\bk)}\right)^2 \dd\bk \\
	&= \lim_{s\rightarrow\infty} \int_{\R^D} \left(\frac{s^{|\bnu|+D} \,g^{(\bnu)}(s\bk)} {\|\bk\|^{-\left(\eta+\frac{D}{2}\right)}}\right)^2 \dd\bk \\
	&= \lim_{s\rightarrow\infty} \int_{\R^D} \frac{s^{2|\bnu|+2D}}{s^{2\eta+D}} \left(\frac{g^{(\bnu)}(\bkk)} {\|\bkk\|^{-\left(\eta+\frac{D}{2}\right)}}\right)^2 \frac{1}{s^D} \dd\bkk \label{substitution} \\
	&= \int_{\R^D} \left(\frac{g^{(\bnu)}(\bkk)} {\|\bkk\|^{-\left(\eta+\frac{D}{2}\right)}}\right)^{\!\!2} \!\dd\bkk \cdot \lim_{s\rightarrow\infty} \frac{s^{2|\bnu|}}{s^{2\eta}} \label{const_int} \\
	&= \begin{cases}
		0 & |\bnu|<\eta \\
		C_{\eta,\bnu} & |\bnu|=\eta \\
		\infty & |\bnu|>\eta
	\end{cases}
\end{align}
where we have abbreviated the finite integral in (\ref{const_int}) with the symbol $C_{\eta,\bnu}$.

\subsection{Constraints on coefficients of Green's functions}
\label{app:GRC}

In this section, we determine which polyharmonic splines have finite $\eta$-norm, i.e.\ we derive a constraint on the Green's functions' coefficients $\bb$.
For readability, we reprint the definitions of the polyharmonic spline $f$ in position and frequency space.
Since the polynomial term does not contribute to the $\eta$-norm, we leave it out.
\begin{align}
	f(\bx) &= \sum_{n=1}^N b_n \|\bx-\bx_n\|^{2\eta} \\
	f(\bk) &= \sum_{n=1}^N b_n \frac{\exp(2\pi i\bk\bx_n)}{\|\bk\|^{2\eta+D}}
\end{align}
We treat the problem in frequency space, where the $\eta$-norm of the polyharmonic spline $f$ is defined as follows, and it is convenient to define the function $h$.
\begin{align}
	|f|_\eta &= \int_{\R^D} \frac{|f(\bk)|^2}{\|\bk\|^{-(2\eta+D)}} \dd\bk \\
	&= \int_{\R^D} \frac{|h(\bk)|^2}{\|\bk\|^{2\eta+D}} \dd\bk \label{k-repr} \\
	h(\bk) &= \|\bk\|^{2\eta+D}\cdot f(\bk) = \sum_{n=1}^N b_n \exp(2\pi i\bk\bx_n)
\end{align}
To understand for which coefficients $\bb$ the norm $|f|_\eta$ is finite, first note that $|h(\bk)|^2$ alternates symmetrically around $0$, and hence the integral (\ref{k-repr}) converges for $\|\bk\|\rightarrow\infty$ for all $2\eta+D>0$, which includes the complete domain of interest $\eta>0$.
The more complicated part is the convergence at the origin $\|\bk\|\rightarrow0$.
Roughly speaking, since the denominator has a zero of degree $2\eta+D$, the numerator must have a zero of degree at least $2\eta$ to achieve convergence.
We will make this precise in the following.

To find coefficients $\bb$ such that $h^2(\bk)$ has a zero of degree $2\eta$, we consider the Taylor expansions of $h(\bk)$ and $h^2(\bk)$.
\begin{align}
	h(\bk) &= \sum_{\bnu} \alpha_{\bnu} \bk^{\bnu} &
	\alpha_{\bnu} &= \frac{\bm{\partial}^{\bnu} h(\bk)}{\bnu!} \label{Taylor} \\
	h^2(\bk) &= \sum_{\bnu} \beta_{\bnu} \bk^{\bnu} &
	\beta_{\bnu} &= \sum_{|\tilde\bnu|\leq|\bnu|} \alpha_{\tilde\bnu} \alpha_{\bnu-\tilde\bnu}
\end{align}
where $\bnu$ and $\tilde\bnu$ are a multi-indices, $\bnu!=\prod_d \nu_d!$, and $|\bnu|=\sum_d \nu_d$.
Note that the Taylor coefficients $\beta_{\bnu}$ vanish up to degree $2\eta$ if and only if $\alpha_{\bnu}$ vanish up to degree $\eta$.
\begin{align}
	\beta_{\bnu} &= 0 \qquad \forall|\bnu|\leq2\eta \label{beta0} \\
	\qquad\Longleftrightarrow\qquad
	\alpha_{\bnu} &= 0 \qquad\forall|\bnu|\leq\eta \label{alpha0}
\end{align}
The condition (\ref{alpha0}) can be written in terms of the coefficients $\bb$ as
\begin{align}
	\alpha_{\bnu} = \frac{\bm{\partial}^{\bnu} h(\bk)}{\bnu!} = \frac{2\pi i\,h(\bk)}{\bnu!} \cdot \sum_n b_n \bx_n^{\bnu} &= 0 \qquad\qquad \forall |\bnu|<\eta \\
	\sum_n b_n\bx_n^{\bnu} &= 0 \qquad\qquad \forall |\bnu|<\eta
\end{align}
which is our sought result.

\paragraph{Convergence of high-order Taylor terms}
We have stated above ``roughly speaking'' that the integral (\ref{k-repr}) $\int\bk^{\bnu}/\|\bk\|^{2\eta+D}\dd\bk$ converges for $|\bnu|>2\eta$.
Now we provide a detailed proof for this.
The idea is to divide the $D$-dimensional integral into an integral over zero-centered hyperspheres and an integral over the radius (or scale) $s$.
Since $\bk^{\bnu}$ scales with $s^{|\bnu|}$, the hypersphere integral of $\bk^{\bnu}$ scales with $s^{|\bnu|+D-1}$.
\begin{align}
	(s\bk)^{\bnu} &= s^{|\bnu|} \bk^{\bnu} \\
	q_{\bnu}(s) &= \int_{\|\bk\|=s} \bk^{\bnu} \dd\bk = s^{|\bnu|+D-1} \,q_{\bnu}(1)
\end{align}
Hence the integral of interest can be expressed as:
\begin{align}
	\int_{\|\bk\|\leq1} \frac{\bk^{\bnu}}{\|\bk\|^{2\eta+D}} \dd\bk
	&= q_{\bnu}(1) \int_0^1 \frac{s^{|\bnu|+D-1}}{s^{2\eta+D}} \dd s \\
	&= \begin{cases}
		\infty & |\bnu| \leq 2\eta \\
		\frac{q_{\bnu}(1)}{|\bnu|-2\eta} & |\bnu| > 2\eta \\
	\end{cases}
\end{align}


\subsection{Inner product of polyharmonic splines}
\label{app:InnerProduct}

In this section, we calculate the inner product $\left<\cdot,\cdot\right>_\eta$ of two polyharmonic splines $f(\bx),\tilde f(\bx)$ with equal nodes $\bx_n$ for half-integer $\eta$.
The splines are specified by the coefficients $\bb,\tilde\bb$ of the Green's functions $g_{\bx_n}$ and the polynomial coefficients $\bm{c},\tilde{\bm{c}}$.
Since the nullspace polynomials do not contribute to the inner product, we neglect them in the following.
\begin{align}
	f(\bx) &= \sum_{n=1}^N b_n\, g_{\bx_n}\!(\bx) + \text{polynomial} \\
	g_{\bx_n}(\bx) &= \|\bx_n\|^{2\eta} \\
	g_{\bx_n}(\bk) &= \tilde C\, \frac{\exp(2\pi i\bk\bx_n)}{\|\bk\|^{2\eta+D}} \label{g_x(k)}
\end{align}
The constant $\tilde C$ stems from the Fourier pair (21) in Sec.\ 3.1 of the main paper.
Since $f$ and $\tilde f$ are sums, we can decompose their inner product as a bilinear form of the coefficients $\bb,\tilde\bb$:
\begin{align}
	\left<f,\tilde f\right>_\eta &= \left<\bb,\tilde\bb\right>_\eta = \sum_{n,m} b_n \tilde b_m \big<g_{\bx_n},g_{\bx_m}\big>_\eta^* = \bb^T H \tilde\bb \label{decomp} \\
	H_{nm} &= \big<g_{\bx_n},g_{\bx_m}\big>_\eta^*
\end{align}
The problem with this decomposition is that single Green's functions $g_{\bx_n}$ have infinite $\eta$-norm, and hence also infinite inner product $\big<g_{\bx_n}, g_{\bx_m}\big>_\eta=\infty$.
The infinities cancel when taking the sum (\ref{decomp}), if the coefficients $\bb,\tilde\bb$ satisfy the constraints (23, 24) of Sec.\ 3.1 of the main paper.
In order to use (\ref{decomp}), however, we need to leave out the infinite terms already in the single components $H_{nm}$, which we have indicated above by adding a star to the inner product $\left<\cdot,\cdot\right>_\eta^*$.
How this is achieve will be explained below.

To determine $H$, we need to calculate the following integral.
\begin{align}
	\big<g_{\bx_n}, g_{\bx_m}\big>_\eta &= \int \frac{g_{\bx_n}\!(\bk) \cdot g^*_{\bx_m}\!(\bk)}{\|\bk\|^{-2\eta-D}} \dd\bk \\
	&= \int \frac{\|\bk\|^{-2\eta-D}\exp(2\pi i\, \bk\bx_n) \cdot \|\bk\|^{-2\eta-D}\exp(-2\pi i\, \bk\bx_m)} {\|\bk\|^{-2\eta-D}} \dd\bk \\
	&= \int \frac{\cos\big(2\pi\bk(\bx_n-\bx_m)\big)} {\|\bk\|^{2\eta+D}} \dd\bk \label{D_int}
\end{align}
The star of $g_{\bx_m}^*$ denotes complex conjugation.
The imaginary part cancels upon integration over $\R^D$ because it has odd symmetry.

To evaluate (\ref{D_int}), we observe that the numerator is constant over all hyperplanes perpendicular to $\bxi=\bx_n-\bx_m$.
To exploit this symmetry, we reparametrize $\bk$ as $[k,\tilde\bk]$ such that $k$ is aligned with $\bxi$, and $\tilde\bk$ contains the $D-1$ components perpendicular to $\bxi$.
\begin{align}
	\xi &= \|\bxi\| = \|\bx_n-\bx_m\| \\
	k &= \bk\cdot\frac{\bxi}{\xi} \\
	\tilde\bk\cdot\bxi &= 0 \\
	\cos\big(2\pi\bk(\bx_n-\bx_m)\big) &= \cos\big(2\pi k\xi\big)
\end{align}
Now we can evaluate the integral (\ref{D_int}) in two steps: first we integrate over the $D-1$ dimensional hyperplane $\tilde\bk$ with fixed $k$, then we integrate over $k$. 

\paragraph{Hyperplane integral}
To integrate (\ref{D_int}) over the $D-1$ dimensions $\tilde\bk$, we observe that the integrand is very similar to the $(D-1)$-dimensional $t$-distribution's pdf.
To obtain the same form, we factor out $1/k^{2\eta+D}$ and then substitute $\tilde\bk$ with $\bkk$ as defined below:
\begin{align}
	\int_{\R^{D-1}} \frac{1}{\|\bk\|^{2\eta+D}} \dd\tilde\bk
	&= \int_{\R^{D-1}}\! \left(k^2 + \|\tilde\bk\|^2\right)^{\!-\big(\eta+\frac{D}{2}\big)} \dd^{D-1}\tilde\bk \\
	&= \frac{1}{k^{2\eta+D}} \int_{\R^{D-1}}\! \left(1 + \frac{\|\tilde\bk\|^2}{k^2}\right)^{\!\!-\big(\eta+\frac{D}{2}\big)} \dd^{D-1}\tilde\bk \\
	&= \frac{1}{\nu^\frac{D-1}{2}\,k^{2\eta+1}}\; \int_{\R^{D-1}}\! \left(1 + \frac{\|\bkk\|^2}{\nu}\right)^{\!\!-\big(\eta+\frac{D}{2}\big)} \dd^{D-1}\bkk \label{D-1_int} \\
	&\hspace{4.5cm} \bkk = \frac{\sqrt{\nu}}{k} \,\tilde\bk \\
	&\hspace{3.6cm} \dd^{D-1}\tilde\bk = \frac{k^{D-1}}{\nu^{(D-1)/2}} \dd^{D-1}\bkk
\end{align}
The above integral is equal to the reciprocal of the multivariate $t$-distribution's normalization factor.
To make the following obvious, we first reprint the result in the $t$-distribution's standard notation, where $p$ denotes the dimension and $\nu$ the number of degrees of freedom, then we replace $(p,\nu)$ with the corresponding expressions in (\ref{D-1_int}).
\begin{align}
	\int_{\R^p} \left(1+\frac{\|\bx\|^2}{\nu}\right)^{-\frac{\nu+p}{2}}\dd\bx &= \frac{\Gamma\big(\frac{\nu}{2}\big)\,(\nu\pi)^{p/2}} {\Gamma\big(\frac{\nu+p}{2}\big)} \\
	p \;&\longrightarrow\; D-1 \\
	\nu \;&\longrightarrow\; 2\eta + 1 \\
	\nu+p \;&\longrightarrow\; 2\eta + D \\
	\frac{1}{\nu^\frac{D-1}{2}} \int_{\R^{D-1}} \left(1+\frac{\|\bkk\|^2}{2\eta+1}\right)^{-\big(\eta+\frac{D}{2}\big)} \dd\bkk
	&= \frac{\Gamma\big(\eta+\frac{1}{2}\big)\,\pi^{\frac{D-1}{2}}} {\Gamma\big(\eta+\frac{D}{2}\big)} \label{hyperplane_result}
\end{align}

\paragraph{Cosine integral}
For the second step of determining (\ref{D_int}) we need to evaluate
\begin{align}
	\int \frac{\cos(2\pi k\xi)}{k^{2\eta+1}} \dd k \label{cos_int}
\end{align}
This integral diverges at $k=0$ for all $\eta>0$.
This divergence can be avoided by neglecting the cosine's Taylor terms up to order $2\eta$, which are exactly the Taylor terms that cancel in $f,\tilde f$ for valid coefficients $\bb,\tilde\bb$, see (\ref{beta0}) of the previous section.

To keep track of the canceling Taylor terms, we will use the functions $\cos_\nu^*$ and $\sin_\nu^*$ in the following, which are defined by
\begin{align}
	\cos_\nu^*(k) &= \cos(k) - \sum_{2i\leq\nu} \frac{(-1)^i}{(2i)!}\,k^{2i} = \sum_{i=\nu+1}^{\infty} \frac{(-1)^i}{(2i)!}\,k^{2i} \label{cos*} \\
	\sin_\nu^*(k) &= \sin(k) - \sum_{2i+1\leq\nu}^\nu \frac{(-1)^i}{(2i+1)!}\,k^{2i+1} = \sum_{i=\nu+1}^{\infty} \frac{(-1)^i}{(2i+1)!}\, k^{2i+1} \label{sin*}
\end{align}

Now we use partial integration with the following variables in order to evaluate integrals of the type (\ref{cos_int}).
\begin{align}
	u_1 &= \cos_\nu^*(ak)                 & u_2 &= \sin_\nu^*(ak)                & v &= \frac{-1}{(j-1)\,k^{j-1}} \\
	\dd u_1 &= -a\sin_{\nu-1}^*(ak) \dd k & \dd u_2 &= a\cos_{\nu-1}^*(ak) \dd k & \dd v &= \frac{1}{k^j} \dd k
\end{align}
\begin{align}
	\int \frac{\cos_\nu^*(ak)} {k^j} \dd k
	&= -\frac{\cos_\nu^*(k)}{(j-1)\,k^{j-1}} - \frac{a}{j-1} \int \frac{\sin_{\nu-1}^*(k)}{k^{j-1}} \dd k \\
	\int \frac{\sin_\nu^*(k)} {k^j} \dd k
	&= -\frac{\sin_\nu^*(k)}{(j-1)\,k^{j-1}} + \frac{a}{j-1} \int \frac{\cos_{\nu-1}^*(k)}{k^{j-1}} \dd k
\end{align}
Repeated application of the above yields the following chain.
Since $\eta$ is a half-integer, $2\eta+1$ is even, and the chain ends with $\sin_1^*(k)=\sin(k)/k$, i.e.\ the well-known sine integral.
\begin{align}
	\int \frac{\cos_{2\eta}^*(k)} {k^{2\eta+1}} \dd k
	&= -\frac{1}{2\eta}\frac{\cos_{2\eta}^*(k)}{k^{2\eta}} + \frac{a}{2\eta(2\eta-1)}\frac{\sin_{2\eta-1}^*(k)}{k^{2\eta-1}} + \dots
	+(-1)^{\etac} \frac{a^{2\eta}}{(2\eta)!}\int \frac{\sin(k)}{k} \dd k \label{chain}
\end{align}
Now we evaluate the above for the limits $0<k<\infty$ and find that all non-integral terms vanish.
For $k=\infty$, we can regard the standard sine and cosine functions (without canceled terms), which are bounded to $[-1,1]$ so that the fractions below vanish at infinity.
For $k=0$, we have to consider the Taylor terms in (\ref{cos*}, \ref{sin*}).
Because all Taylor terms present in $\cos_\nu^*,\sin_\nu^*$ have order larger than $\nu$, they cancel the denominator up to an exponent of $j\leq\nu$.
Together we have
\begin{align}
	\left.\frac{\cos(k)}{k^j}\right|_{k=\infty} = \left.\frac{\sin(k)}{k^j}\right|_{k=\infty} &= 0 \qquad\text{for } j>0 \\
	\left.\frac{\cos_\nu^*(k)}{k^j}\right|_{k=0} = \left.\frac{\sin_\nu^*(k)}{k^j}\right|_{k=0}\;\, &= 0 \qquad\text{for } j\leq\nu
\end{align}
As a result, (\ref{chain}) reduces to a multiple of the sine integral $\int_0^\infty \sin(k)/k\dd k=\pi/2$, and when inserting $a=2\pi\xi$ from (\ref{cos_int}), we obtain:
\begin{align}
	\int \frac{\cos_{2\eta}^*(2\pi\,k\xi)}{k^{2\eta+1}} \dd k = (-1)^{\lceil\eta\rceil} \frac{(2\pi\xi)^{2\eta}}{(2\eta)!} \frac{\pi}{2} \label{cos_int_result}
\end{align}

\paragraph{Putting things together}

Collecting the terms from (\ref{g_x(k)}, \ref{hyperplane_result}, \ref{cos_int_result}), we obtain:
\begin{align}
	H_{nm} = \big<g_{\bx_n}, g_{\bx_m}\big>_\eta^*
	&= \tilde C\, \int \frac{\cos_{2\eta}^*\big(2\pi\bk\bxi\big)} {\|\bk\|^{2\eta+D}} \dd\bk \\
	&= \tilde C\, \frac{\Gamma\big(\eta+\frac{1}{2}\big)\,\pi^\frac{D-1}{2}} {\Gamma\big(\eta+\frac{D}{2}\big)} \int \frac{\cos_{2\eta}^*(2\pi k\xi)}{k^{2\eta+1}} \dd k \\
	&= \tilde C\, \frac{\Gamma\big(\eta+\frac{1}{2}\big)\,\pi^\frac{D-1}{2}} {\Gamma\big(\eta+\frac{D}{2}\big)}\, (-1)^{\lceil\eta\rceil} \frac{(2\pi\xi)^{2\eta}}{(2\eta)!} \frac{\pi}{2} \\
	&= \tilde C\, C_{\eta,D} \,\big\| \bx_n-\bx_m \big\|^{2\eta} \\
	& \qquad C_{\eta,D} = (-1)^{\etac}\frac{(2\pi)^{2\eta}}{2}\, \pi^\frac{D+1}{2}\, \frac{\Gamma\big(\eta+\frac{1}{2}\big)} {\Gamma\big(\eta+\frac{D}{2}\big)}\, \frac{1}{(2\eta)!}
\end{align}
Inserting this expression into (\ref{decomp}) yields the sought inner product.

We conjecture that when replacing the factorial $(2\eta)!$ with $\Gamma(2\eta+1)$, the result is valid for all positive non-integer $\eta$.

\ \\

\subsubsection*{References}

Kwa\'{s}nicki, Mateusz.
\textit{Ten Equivalent Definitions of the Fractional Laplace Operator.}
Fractional Calculus and Applied Analysis, Vol.\ 20, No.\ 1 (2017), pp. 7--51.
\href{https://link.springer.com/10.1515/fca-2017-0002}{doi:10.1515/fca-2017-0002}.
\href{https://arxiv.org/abs/1507.07356}{arXiv:1507.07356}.

\end{document}